# Lattice-Based Fuzzy Medical Expert System for Low Back Pain Management

Debarpita Santra[1*], S. K. Basu[2], J. K. Mondal[1],
and Subrata Goswami[3],

[1] Department of Computer Science & Engineering, University of Kalyani, Kalyani,Nadia, West Bengal-741235,India
[2] Department of Computer Science, Banaras Hindu University, Varanasi 221005, India
[3] ESI Institute of Pain Management, Kolkata 700009, India

{debarpita.cs*, swapankb, jkm.cse, drsgoswami}@gmail.com

**Abstract.** Low Back Pain (LBP) is a common medical condition that deprives many individuals worldwide of their normal routine activities. In the absence of external biomarkers, diagnosis of LBP is quite challenging. It requires dealing with several clinical variables, which have no precisely quantified values. Aiming at the development of a fuzzy medical expert system for LBP management, this research proposes an attractive lattice-based knowledge representation scheme for handling imprecision in knowledge, offering a suitable design methodology for a fuzzy knowledge base and a fuzzy inference system. The fuzzy knowledge base is constructed in modular fashion, with each module capturing interrelated medical knowledge about the relevant clinical history, clinical examinations and laboratory investigation results. This approach in design ensures optimality, consistency and preciseness in the knowledge base and scalability. The fuzzy inference system, which uses the Mamdani method, adopts the triangular membership function for fuzzification and the Centroid of Area technique for defuzzification. A prototype of this system has been built using the knowledge extracted from the domain expert physicians. The inference of the system against a few available patient records at the ESI Hospital, Sealdah has been checked. It was found to be acceptable by the verifying medical experts.

## 1 Introduction

India is confronting a huge void in its healthcare infrastructure, which can be filled with enormous application of information technology in the existing solutions. Medical expert system [1], a computer based new-generation medical decision support tool powered by Artificial Intelligence, bridges the gap between medical and computer sciences by engaging doctors and computer scientists to work in a collaborative manner so that the clinical parameters, symptoms, knowledge of the expert doctors can be integrated in one system.

This paper deals with developing a medical expert system for Low Back Pain (LBP) [2] management. LBP is a common medical problem that deprives many individuals in our country of leading their normal routine lifestyles. Diagnosing LBP is challenging, because it requires highly specialized knowledge involving a complex anatomical and physiological structure of a human being as well as diverse clinical considerations. With no standard techniques accepted and widely-used in clinical practice for measuring pain accurately, it is quite difficult to describe pain in terms of crisp quantifiable attribute-values. Also, how a person feels pain depends on his/her physiological, psychological, or other environmental factors. That's why the pain intensity of an individual may differ from the other. These kinds of clinical uncertainties can be resolved with the help of fuzzy logic and other mathematical techniques.

While fuzzy logic deals with imprecision in knowledge, the properties like completeness and non-redundancy in the knowledge base are ensured by the application of lattice theory in the design. In this paper, a lattice-based fuzzy medical expert system for LBP diagnosis has been proposed. The triangular membership functions have been used for fuzzifying the linguistic values which are used to describe the clinical attributes, and the Mamdani inference approach has been used for designing the inference engine. The system has been successfully tested with twenty LBP patient records at the ESI Hospital Sealdah, West Bengal.

The rest of the paper is organized as: section 2 provides an outline of related works. In section 3, design issues of the intended medical expert system have been addressed. In section 4, the proposed system has been illustrated with a simple example from LBP domain. Finally, section 5 concludes the paper.

## 2 Related Works

Development of medical expert systems began in early 70's with MYCIN [3], followed by CASNET [4], INTERNIST [5], ONCOCIN [6], PUFF [7], etc. As uncertainty pervades in almost all the stages of clinical decision making, some existing medical expert systems used different mathematical techniques to handle inconsistency, imprecision, incompleteness, or other issues. While Rough Set theory [8] is widely used for dealing with inconsistency, and incompleteness in knowledge, fuzzy logic [9] is extensively used for handling imprecision in knowledge. A number of fuzzy medical expert systems have been reported in the literature for diagnosis of heart disease, typhoid fever, abdominal pain and for providing clinical support in intensive care unit.

A fuzzy medical expert system for heart disease diagnosis was proposed based on the databases available in V.A. Medical Centre, Long Beach, and Cleveland Clinic databases. The system took help of Mamdani inference method with thirteen input fields, and one output field [10]. The decision support system for the diagnosis of typhoid fever used triangular membership function for each of the decision variables, and also Mamdani fuzzy inference method for designing the inference system. The proposed system was successfully evaluated with the typhoid patient cases from Federal Medical Centre, Owo, Ondo State-Nigeria [11]. The fuzzy expert system proposed for medical diagnosis of acute abdominal pain, considered around 200 symptoms. Knowledge about these symptoms was stored in the knowledge base in the form of about 60 rules and 4000 fuzzy relations. The partial test of the system was done with 100 patients at the 'Chirurgische Klinik' in Dusseldorf [12].

## 3  Design of Proposed System

Development of a fuzzy medical expert system for LBP diagnosis requires efficient design of a fuzzy knowledge base, and a fuzzy inference system. The block diagram of the intended system is shown in figure 1.

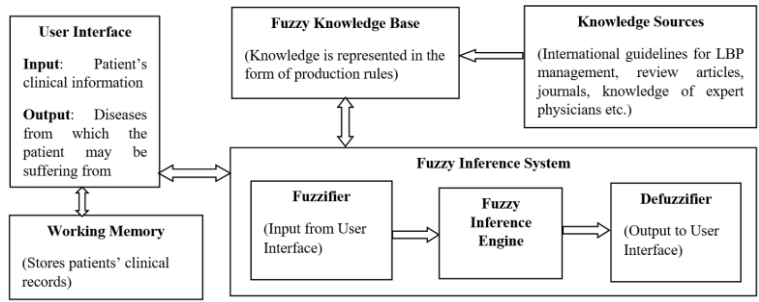

Fig 1: Block diagram of the intended fuzzy medical expert system for LBP diagnosis

When an LBP patient comes for treatment, all the significant clinical information (clinical history, reports of local/general examinations, results of clinical investigation etc.) of the patient are fed into the system through the user interface. The fuzzy inference system accepts the inputted patient information, matches them with the knowledge stored in the fuzzy knowledge base, and concludes with a list of diseases from which the patient may be suffering from.

### 3.1   Design of Fuzzy Knowledge Base

We assume that there is a finite set $D$ of $x$ LBP diseases in the literature. Assessment of different LBP diseases goes through three different phases: *phase* 1 for finding a list of probable diseases based on the relevant clinical history of a patient $X$; *phase* 2 for determining a list of probable diseases based on the local/general examinations of $X$; and *phase* 3 for finding a list of probable diseases based on the clinical investigation results. Considering that there is a finite non-empty set $A$ of $n$ clinical parameters / attributes relevant for assessment of all the LBP diseases in the literature, $n_1$ ($0 < n_1 < n$) clinical attributes comprising the set $A_1$ ($\subseteq A$) are used for collection of clinical history in *phase* 1, $n_2$ ($0 < n_2 < n$) attributes forming the set $A_2$ ($\subseteq A$) are there for acquiring the local/general examination reports of LBP patients in *phase* 2, and $n_3$ ($0 < n_3 < n$) attributes constituting the set $A_3$ ($\subseteq A$) are considered for capturing the clinical investigation results of LBP patients in *phase* 3. Here, $A_1$, $A_2$, $A_3$ are disjoint subsets of *A*.

With no standard techniques accepted and widely-used in clinical practice for measuring pain accurately, it is quite difficult to describe pain in terms of crisp quantifiable attribute-values. Use of Numeric Rating Scale (NRS) as shown in figure 2 for measuring pain intensity is a popular approach.

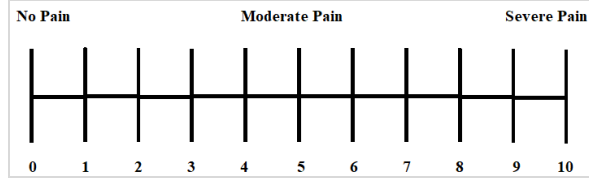

Fig 2: Numeric Rating Scale for pain assessment

In NRS, the three pain intensity values 'No Pain', 'Moderate Pain', and 'Severe Pain' cannot be crisply defined; rather, there exists imprecision in defining the range for every linguistic value.

The fuzzy knowledge base for the proposed medical expert system holds knowledge related to *x* LBP diseases, where each disease is characterized by the clinical attributes belonging to *A*. Every piece of knowledge needs to be represented in the form of production rules. The rules are generated from the acquired knowledge in a systematic fashion with the help of a lattice structure. The lattice structure actually works as the fuzzy knowledge base.

*Rule Generation using Lattice*

Prior to formation of a lattice structure, a fuzzy information system $S_f$ is designed using the acquired knowledge, which contains $M$ ( > 0) rows and $N$ ( = $n$ + 2) columns. In $S_f$, each tuple is represented as < ($d$, $v_f(d)$), {($a_j$, $v_f(a_j)$) | $a_j \in A$ ($1 \leq j \leq n$) is the $j^{th}$ clinical attribute for the disease $d$, and $v_f(a_j)$ denotes the linguistic value of $a_j$}, $r_s$ >, where $d$ represents an LBP disease, $v_f(d)$ represents the linguistic value associated with the chance of occurrence of $d$, and $r_s$ denotes the reliability strength of a production rule of the form [$\cup_{j=1 \text{ to } n}$ {($a_j$, $v_f(a_j)$)} $\rightarrow$ ($d$, $v_f(d)$)]. The reliability strength $r_s$ ($0 \leq r_s \leq 1$), which measures how much reliable the piece of knowledge is, is determined based on the nature of knowledge sources used for acquiring the knowledge. For example, if a piece of knowledge is acquired from an international guideline, the reliability strength for the piece of knowledge is very high, compared to that of another piece of knowledge acquired from a case study or expert consensus.

The fuzzy information system $S_f$ can formally be represented as $S_f = (D_f, A_f, R_f)$, where $D_f$ is a set of (disease, linguistic_value) pairs, $A_f$ is a set of (attribute, linguistic_value) pairs, and $R_f$ is the set of corresponding reliability strengths. For every subset $B_f$ of $A_f$, a binary indiscernibility relation $I(B_f)$ [8] on $D_f$ is defined as follows: two different (decision, linguistic_value) pairs ($d_i$, $v_f(d_i)$) and ($d_j$, $v_f(d_j)$) [$i \neq j$] are indiscernible by the subset of attributes $B_f$ in $A_f$, if the linguistic value corresponding to an attribute $b_f \in B_f$ is the same for both the pairs ($d_i$, $v_f(d_i)$) and ($d_j$,

$v_f(d_j)$). The equivalence class of the relation $I(B_f)$ is called an elementary set in $B_f$ [8]. The construction of equivalence classes involving the elements in $A_f$ and $D_f$ is performed in a systematic fashion, as given below in a procedure called *Equivalence_Class_Construction*( ). The algorithm uses the fuzzy information system $S_f$ as input.

**Procedure** *Equivalence_Class_Construction* ($S_f$)
Begin
   $k_1 = 0$       // $k_1$ is a temporary variable
   $E_f = \varnothing$      // $E_f$ is the set of elementary sets, which is initially empty
  While $k_1 \leq n$ do //$n$ is the no. of clinical attributes
    For $k_2 = 1$ to $^nC_{k_1}$ do    // $k_2$ is a temporary variable
      $B_f =$ **Combination**$(A_f, k_1)$  // The procedure **Combination**$(A_f, k_1)$ returns a
                                     set of $k_1$ clinical attributes from a set of $n$
                                       attributes in each iteration
      $E_f = E_f \cup [D_f/B_f]$    // $[D_f/B_f]$ denotes the elementary sets of $D_f$ determined by $B_f$
    End for
    $k_1 = k_1 + 1$
  End while
Return $E_f$
**End** *Equivalence_Class_Construction*

If a maximum of $K$ instructions is executed by the procedure **Combination**() in *Equivalence_Class_Construction*() procedure, the total no. of operations performed by the latter is ($\sum_{k_1=0 \text{ to } n} K \times k_1 \times {}^nC_{k_1}$).

All the knowledge contained in the fuzzy information system $S_f$ is stored efficiently in the fuzzy knowledge base $KB_f$. Initially, $KB_f = \varnothing$. Every time a new clinical attribute is encountered, information regarding the attribute is added to the existing $KB_f$ using 'set union' operation. With $n$ ($> 0$) clinical attributes in $A$, the knowledge base is modified as $KB_f = KB_f \cup \{a_i \mid a_i \ (\in A)$ is a clinical attribute, where $1 \leq i \leq n\}$. To extract detailed knowledge about each of the $n$ elements in $KB_f$, the attributes are combined with each other in an orderly fashion. First, $^nC_2$ combinations are made taking exactly two different attributes at a time from the $n$ attributes. In this case, $KB_f = KB_f \cup \{(a_i, a_j) \mid a_i$ and $a_j$ are the two clinical attributes in $A$ where $a_i \neq a_j$ and $1 \leq i, j \leq n\}$. In the same way, $^nC_k$ combinations of attributes are found, where $2 \leq k \leq n$, and the knowledge base is updated accordingly. Now the $KB_f$ holds a total of $(2^n - 1)$ attribute combinations. So, a set $S$ constructed using all the attribute combinations in $KB_f$ acts as the power set of $A$, and is represented as $S = \{\phi, \{a_1\}, \{a_2\}, \{a_3\},\ldots,\{a_n\}, \{a_1, a_2\}, \{a_1, a_3\},\ldots,\{a_{n-1}, a_n\}, \{a_1, a_2, a_3\},\ldots,\{a_{n-2}, a_{n-1}, a_n\},\ldots\{a_1, a_2,\ldots,a_{n-1}, a_n\}\}$.

Now a relation $R$ is considered, which acts as subset equality i.e. $\subseteq$. Suppose $\alpha, \beta, \gamma \subseteq S$. So, $[S; \subseteq]$ is a poset as
i) $\forall \alpha \in S : (\alpha \subseteq \alpha)$ (holds reflexivity relationship)
ii) $\forall (\alpha, \beta) \in S : \alpha \subseteq \beta$ and $\beta \subseteq \alpha \Rightarrow (\alpha = \beta)$ (holds antisymmetry relationship)
iii) $\forall (\alpha, \beta, \gamma) \in S : \alpha \subseteq \beta, \beta \subseteq \gamma \Rightarrow (\alpha \subseteq \gamma)$ (holds transitivity relationship)

Now, two elements $x^\square$ ($\in S$) and $y^\square$ ($\in S$) are taken to construct $\{x^\square, y^\square\}$. This set would always have a *glb* [greatest lower bound] as $(x^\square \cap y^\square)$ ($\in S$) and *lub* [least upper bound] as $(x^\square \cup y^\square)$ $(\in S)$. This is true for any $x^\square$ $(\in S)$ and $y^\square$ $(\in S)$. So, $S$ is a lattice of order $n$. Each node in the lattice contains three types of information: a set $B_f$ of clinical attributes associated with their corresponding linguistic values as recorded in $S_f$, the elementary sets obtained against $B_f$ using the procedure *Equivalence_Class_Construction()*, and the reliability strength against each element belonging to the elementary sets.

Suppose, a node in the lattice $S$ contains $m$ ( $> 0$) elementary sets $E_1, E_2,\ldots, E_m$ against the set of attributes $B_f$. Assume that $B_f$ contains $k$ ( $> 0$) clinical attributes $\{a_i, a_{i+1}, a_{i+2}, \ldots, a_{i+k}\}$ with each attribute $a_{k1}$ ($i \leq k1 \leq (i+k)$) of $B_f$ has a linguistic value $v_{k1}$. The corresponding elementary set $E_j$ ($1 \leq j \leq m$) contains $l$ ( $> 0$) ($d$, $v_j(d)$) pairs, where $d$ denotes an LBP disease and $v_j(d)$ denotes the linguistic value associated with the chance of occurrence of the disease $d$. A maximum of $l$ production rules would be generated of the form $[(a_i, v_i) \text{ AND } (a_{i+1}, v_{i+1}) \text{ AND } \ldots \text{ AND } (a_{i+k}, v_{i+k}) \rightarrow E_j]$. If all the $l$ ($\leq x$) elements in $E_j$ contains $l$ different LBP diseases, then the total number of production rules generated against $B_f$ is $l$. If on the other hand, at least two elements in $E_j$ contain the same disease with two distinct linguistic values $val_1$ and $val_2$ respectively, the issue is resolved with the help of associated reliability strengths. An item with greater reliability strength obviously gets priority over the other elements. The same reliability strength for more than one such conflicting items leads to incorporation of another mathematical technique(s) for handling inconsistency in knowledge. Handling of inconsistencies is out of scope for this paper.

### 3.2 Design of Fuzzy Inference System

The fuzzy inference system consists of a fuzzifier, an inference engine, and a defuzzifier. The fuzzifier fuzzifies the raw linguistic values associated with the clinical attributes in $A$ using triangular membership function. Corresponding to the set $A$, a fuzzy set $A_F$ is first defined as $A_F = \{(a_i, \mu(a_i))| a_i \in A, \mu(a_i) \in [0, 1]\}$, where $\mu(a_i)$ denotes the membership function of $a_i$.

Any inference engine of a medical expert system performs two tasks: (i) it matches the inputted patient information with the stored knowledge; (ii) the matched knowledge is processed for making reliable diagnostic conclusions. Use of a lattice structure as the knowledge base makes the matching process easier. As the information about an LBP patient may include only $p$ ($\leq n$) clinical attributes, it would be sufficient only to search the nodes at the $p^{th}$ level of the lattice $S$. This kind of matching strategy reduces the search time in the knowledge base to a great extent. The fuzzy inference engine uses the Mamdani approach [11], and the defuzzification process uses the Centroid of Area method [11].

As LBP diagnosis goes mainly through three phases, three different lattice structures are formed to store knowledge. The inference engine is also executed phase-wise, and the three different lists of probable diseases are obtained from these phases. In each phase, the list of probable diseases gets refined. The list of diseases obtained after *phase* 3 would be regarded as the final outcomes of the medical expert system. This type of modular approach ensures scalability in the design.

## 4 Illustration and Implementation

For the sake of simplicity, this paper considers only five important clinical history parameters namely 'pain at low back area' ($a_1$), 'pain at legs' ($a_2$), 'pain at rest' ($a_3$), 'pain during forward bending' ($a_4$), and 'pain during backward bending' ($a_5$) as input. No clinical examination and investigation parameters have been used in this paper. Only three linguistic values 'No', 'Moderate', and 'Severe' are considered as the values for the set $A_1$ of attributes $\{a_1, a_2, a_3, a_4, a_5\}$, and the ranges of these fuzzy values are defined as [0, 4], [3, 7], and [6, 10], respectively. The membership function graph for these linguistic variables is shown in figure 3(a).

The designed medical expert system outputs the chances of occurrence for only five major LBP diseases namely 'Sacroiliac Joint Arthropathy' ($d_1$), 'Facet Joint Arthropathy' ($d_2$), 'Discogenic Pain' ($d_3$), 'Prolapsed Intervertebral Disc Disease' ($d_4$), and 'Myofascial Pain Syndrome' ($d_5$). Chance of occurrence of every disease can be described by four linguistic values 'No', 'Low', 'Moderate', and 'High', and the corresponding ranges of values are defined as [0, 10], [8, 25], [20, 70], and [60, 100], respectively. The membership function graph corresponding to the chance of occurrence of the LBP diseases is shown in figure 3(b).

Using the acquired knowledge, an information system $S_f$ is first constructed involving the clinical attributes in $A_1$, as shown in table 1. Using the attributes in $A_1$, a lattice structure is constructed as shown in figure 4. Against each node in the lattice, the elementary sets are obtained using the procedure **Equivalence_Class_Construction**() and the information system $S_f$. In this way, the fuzzy knowledge base is constructed.

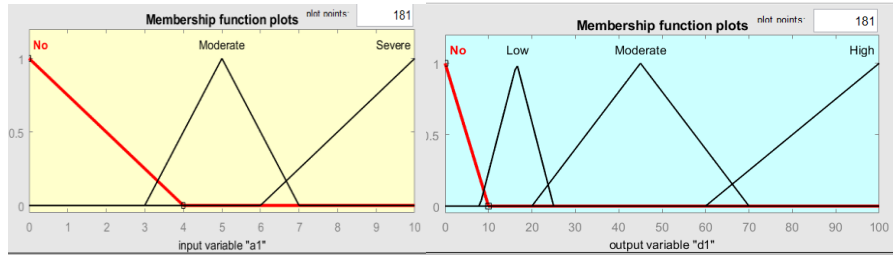

**Fig 3(a)**: Membership function for $a_1$  **Fig 3(b)**: Membership function for $d_1$

Table 1: Information system $S_f$ corresponding to clinical attributes in $A_1$

| (disease, linguistic value) | $a_1$ | $a_2$ | $a_3$ | $a_4$ | $a_5$ | Reliability strength ($r_s$) |
|---|---|---|---|---|---|---|
| ($d_1$, High) | Moderate | No | Severe | Moderate | Moderate | 0.8 |
| ($d_2$, High) | Moderate | No | No | No | Severe | 0.7 |
| ($d_3$, Moderate) | Moderate | No | No | No | No | 0.6 |
| ($d_4$, High) | No | Severe | No | Severe | No | 0.6 |
| ($d_5$, Low) | Moderate | No | Moderate | No | No | 0.4 |

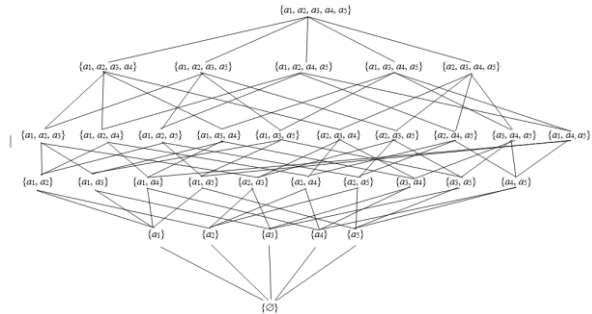

**Fig 4**: Lattice structure formed using the attributes in $A_1$

For the proposed fuzzy expert system, the surface viewers of some fields are shown in figure 5.

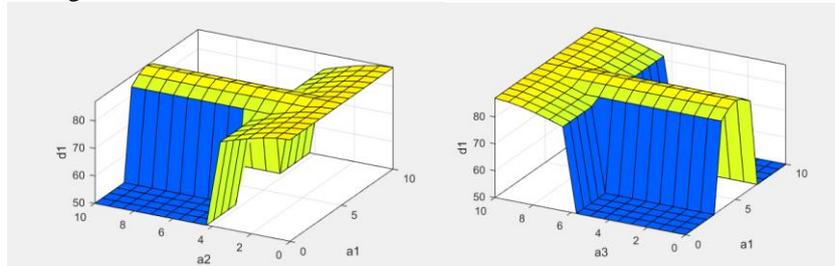

**Fig 5**: Surface viewers for disease $d_1$ against the attributes $a_1$, $a_2$, and $a_3$

A prototype of the designed system has been tested with 20 LBP patient cases obtained from the ESI Hospital, Sealdah, West Bengal. For 18 cases, the achieved results have matched with the expected results. A sample test case has been taken with input values for $a_1 = 4.8$, $a_2 = 3.98$, $a_3 = 2.1$, $a_4 = 5$, and $a_5 = 1.94$. The partial-snapshot of the rule-viewer for this scenario is shown in figure 6. The outputs from the rule-viewer say that the chance of occurrence of the diseases $d_1$, $d_2$, and $d_4$ are high, while that of $d_3$ is moderate, and that of $d_5$ is low.

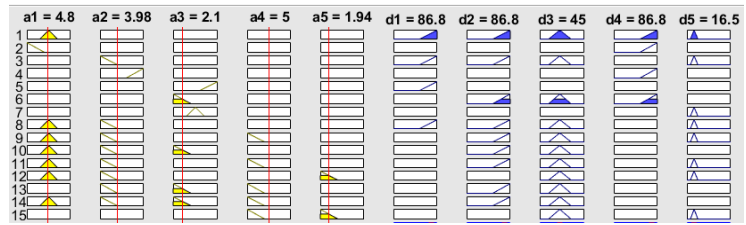

**Fig 6**: Partial snapshot of the rule viewer for the sample test case

## 5 Conclusion

The advantages of developing a lattice-based fuzzy medical expert system are: system completeness, non-redundancy, and preciseness in the knowledge stored in the knowledge base. As reliability is a major concern in medical expert systems, the proposed knowledge base design methodology and inferencing strategy act as a firm basis for development of the intended expert system for LBP management. As the number of LBP diseases is upper bounded by 10 as per expert knowledge, the time and space complexity would not be of much concern. The proposed schemes can be easily extended for design of a full-fledged medical expert system for LBP management in future.